\documentclass[runningheads,a4paper]{llncs}

\usepackage{amssymb}
\usepackage{graphicx}
\usepackage{amsmath}
\usepackage{tabulary}
\usepackage{multirow}
\usepackage{booktabs}  
\usepackage{caption}
\usepackage{enumerate}
\usepackage{array}
\usepackage{booktabs}
\usepackage{url}
\usepackage[pagebackref=false,breaklinks=true,letterpaper=true,colorlinks,bookmarks=false]{hyperref}

\usepackage{url}
\urldef{\mailsa}\path|{jianzhu.guo, xiangyu.zhu, zlei, szli}@nlpr.ia.ac.cn|

\newcommand{\keywords}[1]{\par\addvspace\baselineskip
\noindent\keywordname\enspace\ignorespaces#1}

\newcommand*{\rom}[1]{\uppercase\expandafter{\romannumeral #1}}

\begin{document}

\mainmatter  

\title{Face Synthesis for Eyeglass-Robust Face Recognition}


%
%
\author{Jianzhu Guo, Xiangyu Zhu\thanks{Corresponding author}, Zhen Lei
\and Stan Z. Li
}
%

\institute{CBSR\&NLPR, Institute of Automation, Chinese Academy of Sciences, Beijing, China\\
University of Chinese Academy of Sciences, Beijing, China\\
\mailsa\\
}

%
%

\maketitle

\begin{abstract}
In the application of face recognition, eyeglasses could significantly degrade the recognition accuracy. A feasible method is to collect large-scale face images with eyeglasses for training deep learning methods. However, it is difficult to collect the images with and without glasses of the same identity, so that it is difficult to optimize the intra-variations caused by eyeglasses. In this paper, we propose to address this problem in a virtual synthesis manner. The high-fidelity face images with eyeglasses are synthesized based on 3D face model and 3D eyeglasses.
Models based on deep learning methods are then trained on the synthesized eyeglass face dataset, achieving better performance than previous ones. Experiments on the real face database validate the effectiveness of our synthesized data for improving eyeglass face recognition performance. An eyeglass face dataset named MeGlass is made public at \url{https://github.com/cleardusk/MeGlass}.

\end{abstract}

\section{Introduction}
\label{sec:introduction}
In recent years, deep learning based face recognition systems~\cite{taigman2014deepface,sun2014deep,schroff2015facenet,liu2017sphereface,guo2020learning} have achieved great success, such as Labeled Faces in the Wild (LFW)~\cite{huang2007labeled}, YouTube Faces DB (YFD)~\cite{wolf2011face}, and MegaFace~\cite{kemelmacher2016megaface}.

However, in practical applications, there are still extra factors affecting the face recognition performance, e.g., facial expression, poses, occlusions etc. Eyeglasses, especially black-framed eyeglasses significantly degrade the face recognition accuracy (see Table~\ref{tab:resnet22_A}). There are three common categories of eyeglasses: thin eyeglasses, thick eyeglasses, and sunglasses. In this work, we mainly focus on the category of thick black-framed eyeglasses, since the effects of thin eyeglasses are tiny, while the impact of sunglasses are too high because of serious identity information loss in face texture.

The main contributions of this work include:
1) A eyeglass face dataset named MeGlass, including about $1.7K$ identities, is collected and cleaned for eyeglass face recognition evaluation.
2) A virtual eyeglass face image synthesis method is proposed. An eyeglass face training database named MsCeleb-Eyeglass is generated, which helps improve the robustness to eyeglass.
3) A novel metric learning method is proposed to further improves the face recognition performance, which is designed to adequately utilize the synthetic training data.

The rest of this paper is organized as follows. Section~\ref{sec:related_work} reviews several related works. Our proposed methods are described in Section~\ref{sec:proposed_method}. The dataset description is in Section~\ref{sec:dataset_description}. Extensive experiments are conducted in Section~\ref{sec:experiments} to validate the effectiveness of our synthetic training data and loss function. Section~\ref{sec:conclusion} summarizes this paper.

\section{Related Work}
\label{sec:related_work}
\quad \ \ \textbf{Automatic eyeglasses removal.} 
Eyeglasses removal is another method to reduce the effect of eyeglasses on face recognition accuracy. Several previous works~\cite{saito1999estimation,wu2004automatic,du2005eyeglasses,park2005glasses} have studied on automatic eyeglasses removal. Saito et al.~\cite{saito1999estimation} constructed a non-eyeglasses PCA subspace using a group of face images without eyeglasses, one new face image was then projected on it to remove eyeglasses. Chenyu Wu et al.~\cite{wu2004automatic} proposed an intelligent image editing and face synthesis system for automatic eyeglasses removal, in which eyeglasses region was first detected and localized, then the corrupted region was synthesized adopting a statistical analysis and synthesis approach.
Park et al.~\cite{park2005glasses} proposed a recursive process of PCA reconstruction and error compensation to further eliminate the traces of thick eyeglasses.
However, these works did not study the quantitative effects of eyeglasses removal on face recognition performance. 

\textbf{Virtual try-on.}
Eyeglass face image synthesis is similar to virtual eyeglass try-on. Recently, eyeglasses try-on has drawn attentions in academic community. Niswar et al.~\cite{niswar2011virtual} first reconstructed 3D head model from single image, 3D eyeglasses were next fitted on it, but it lacked the rendering and blending process compared with our synthesis method. Yuan X et al.~\cite{yuan2017magic} proposed a interactive real time virtual 3D eyeglasses try-on system.
Zhang, Q et al.~\cite{zhang2017virtual} firstly took the refraction effect of corrective lenses into consideration. They presented a system for trying on prescription eyeglasses, which could produce a more real look of wearing eyeglasses.

\textbf{Synthetic images for training.}
Recently, synthetic images generated from 3D models have been studied in computer vision~\cite{su2015render,massa2016deep,stark2010back,liebelt2010multi}. These works adopted 3D models to render images for training object detectors and viewpoint classifiers. Because of the limited number of 3D models, they tweaked the rendering parameters to generate more synthetic samples to maximize the model usage.
\section{Proposed Method}
\label{sec:proposed_method}
\subsection{Eyeglass image synthesis}
\label{sec:eyeglass_image_synthesis}
\begin{figure*}[t]
    \centering
    \includegraphics[width=0.95\linewidth]{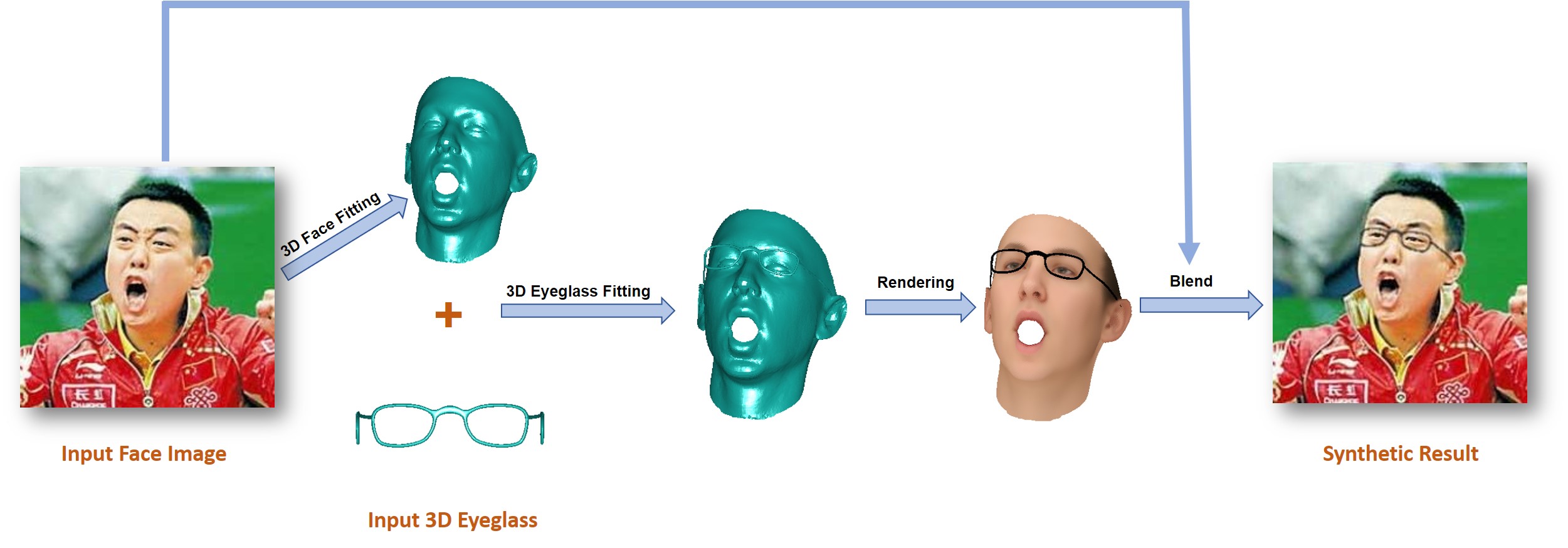}
    \vspace{-1em}
    \caption{The pipeline of eyeglass synthesis.}
    \label{fig:add_glass_pipeline}
\end{figure*}

We describe the details of eyeglass face synthesis in this section.
To generate faces with eyeglasses, we estimate the positions of the 3D eyeglasses based on the fitted 3D face model and then render the 3D eyeglasses on the original face images. The whole pipeline of our eyeglass faces synthesis is shown in Fig.~\ref{fig:add_glass_pipeline}.
Firstly, we reconstruct the 3D face model based on 3DMM fitting or regression methods~\cite{zhu2015high,3ddfa_cleardusk,guo2020towards}, which is robust to pose. 
Secondly, the 3D eyeglass is fitted on the reconstructed 3D face model. The fitting is based on the corresponding anchor points on the 3D eyeglass and 3D fitted face model, where the indices of these anchor points are annotated beforehand. Then z-buffer algorithm and Phong illumination model are adopted for rendering, and the rendered eyeglass image is blended on the original image to generate the final synthetic result.
\par The 3D eyeglass fitting problem is formed as Eq.~\ref{eq:eyeglass_fit}, where $f$ is the scale factor, $Pr$ is the orthographic projection matrix, $p_g$ is the anchor points on 3D eyeglass, $p_f$ is the anchor points on reconstructed 3D face model, $R$ is the $3 \times 3$ rotation matrix determined by pitch($\alpha$), yaw($\beta$), and roll($\gamma$) and $t_{3d}$ is the translation vector.

\begin{figure}[t]
    \centering
    \includegraphics[width=0.95\linewidth]{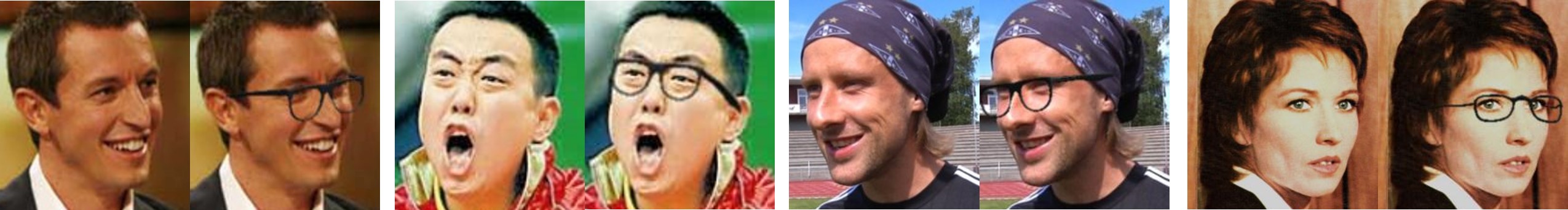}
    \vspace{-0.8em}
    \caption{Four pairs of origin-synthesis images selected from MsCeleb.}
    \label{fig:virtual_black_eyeglasses}
\end{figure}

\begin{equation}
	\label{eq:eyeglass_fit}
    \arg \min_{f, Pr, R, t_{3d}} ||f * Pr * R * (p_g + t_{3d}) - p_f|| ,
\end{equation}

Although the amount of images of MsCeleb is large, the model may overfit during training if the patterns of synthetic eyeglasses are simple. To increase the diversity of our synthetic eyeglass face images, we inject randomness into two steps of our pipeline: 3D eyeglass preparation and rendering.
For 3D eyeglasses, we prepare four kinds of eyeglasses with different shapes and randomly select one as input. For eyeglass rendering, we explore three sets of parameters: light condition, pitch angle and vertical transition of eyeglass. 
For the light condition, the energies and directions are randomly sampled. Furthermore, to simulate the real situations of eyeglass wearing, we add small perturbations to the pitch angle ($[-1.5, 0.8]$) and vertical transition ($[1, 2]$ pixel).
Finally, we put together the synthetic eyeglass face images with original images as our training datasets.


\subsection{Network and loss}
\label{sec:network_loss}
\subsubsection{Network.}
We adapt a $22$ layers residual network architecture based on~\cite{he2016deep} to fit our task. The original ResNet is designed for ImageNet~\cite{deng2009imagenet}, the input image size is $224 \times 224$, while ours is $120 \times 120$. Therefore, we substitute the original $7 \times 7$ convolution in first layer with $5 \times 5$ and stack one $3 \times 3$ convolution layer to preserve dimensions of feature maps.
The details of our ResNet-22 are summarized in Table~\ref{tab:resnet22}.

\begin{table}[t]
    \caption{Our ResNet-22 network structure. Conv3.x, Conv4.x and Conv5.x indicates convolution units which may contain multiple convolution layers and residual blocks are shown in double-column brackets. E.g., [3 $\times$ 3, 128] $\times$ 3 denotes 3 cascaded convolution layers 128 feature maps with filters of size 3 $\times$ 3, and S2 denotes stride 2. The last layer is global pooling.}
	\centering
	
      \begin{tabular}{lc}
      \noalign{\smallskip}\hline\noalign{\smallskip}
      \textbf{Layers} & \textbf{22-layer CNN} \\
      \noalign{\smallskip}\hline\noalign{\smallskip}
      \specialrule{0em}{1pt}{1pt}
      Conv1.x & [5$\times$5, 32]$\times$1, S2 \\\hline
      \specialrule{0em}{1pt}{1pt}
      Conv2.x & [3$\times$3, 64]$\times$1, S1 \\\hline
      \specialrule{0em}{3pt}{1pt}
      Conv3.x & $\left[\begin{aligned}&3\times3, 128\\&3\times3, 128\end{aligned}\right]\times 3$, S2 \\\hline
      \specialrule{0em}{3pt}{1pt}
      Conv4.x & $\left[\begin{aligned}&3\times3, 256\\&3\times3, 226\end{aligned}\right]\times 4$, S2 \\\hline
      \specialrule{0em}{3pt}{1pt}
      Conv5.x & $\left[\begin{aligned}&3\times3, 512\\&3\times3, 512\end{aligned}\right]\times 3$, S2 \\\hline
      \specialrule{0em}{3pt}{1pt}
      Global Pooling & 512 \\
      \noalign{\smallskip}\hline
      \end{tabular}
      
      \label{tab:resnet22}
\end{table}

\subsubsection{Loss.} Due to the disturbance of eyeglass on feature discrimination, we propose the Mining-Contrasive loss based on~\cite{sun2014deep2} to further enlarge the inter-identity differences and reduce intra-identity variations.
The form of our proposed loss is in Eq.~\ref{eq:mining_contrasive}.\begin{equation}
	\label{eq:mining_contrasive}
	L_{mc} = -\frac{1}{2|\mathcal{P}|} \sum_{(i,j) \in \mathcal{P}} d(f_i,f_j) + \frac{1}{2|\mathcal{N}|} \sum_{(i,j) \in \mathcal{N}} d(f_i,f_j) .
\end{equation}

Where $f_i$ and $f_j$ are vectors extracted from two input image samples, $\mathcal{P}$ is hard positive samples set, $\mathcal{N}$ is hard negative samples set, 
$d(f_i, f_j) = \frac{f_i \cdot f_j}{ ||f_i||_2 ||f_j||_2 }$ is cosine similarity between extracted vectors.

\textbf{Gradual sampling}. 
Besides, we employ the gradual process into data sampling to make the model fit the synthetic training images in a gentle manner.
In naive sampling, the probability of eyeglass face image of each identity is fixed at $0.5$. It means that we just brutally mix MsCeleb and MsCeleb-Eyeglass datasets.
We then generalize the sampling probability as $p = \lambda \cdot n + p_0$, where $n$ is the number of iterations, $\lambda$ is the slope coefficient determining the gradual process, $p_0$ is the initialized probability value.
\section{Dataset Description} 
\label{sec:dataset_description}

In this section, we describe our dataset in detail and the summary is shown in Table~\ref{tab:dataset_summary}.

\begin{table*}[t]
	\caption{Summary of dataset description. G and NG indicate eyeglass and non-eyeglass respectively. Mixture means the MsCeleb and MsCeleb-Eyeglass.}
    \centering
    \begin{tabular}{lcccc}
        \noalign{\smallskip}\hline\noalign{\smallskip}
            \textbf{Dataset} & \textbf{Identity} & \textbf{Images} & \textbf{G} & \textbf{NG} \\
            \noalign{\smallskip}\hline\noalign{\smallskip}
            MeGlass & $1,710$ & $47,917$ & $14,832$ & $33,085$\\
            Testing set & $1,710$ & $6,840$ & $3,420$ & $3,420$ \\
            Training set(MsCeleb) & $78,765$ & $5,001,877$ & - & -\\
            Training set(Mixture)& $78,765$ & $10,003,754$ & - & -\\
        \noalign{\smallskip}\hline
    \end{tabular}
    
    \label{tab:dataset_summary}
\end{table*}

\subsection{Testing set}

We select real face images with eyeglass from MegaFace~\cite{kemelmacher2016megaface} to form the MeGlass dataset.
We first apply an attribute classifier to classify the eyeglass and non-eyeglass face images automatically. After that, we select the required face images manually from the attibute-labeled face images. Our MeGlass dataset contains $14,832$ face images with eyeglasses and $33,087$ images without eyeglasses, from $1,710$ subjects.

\begin{figure*}[t]
    \centering
    \includegraphics[width=0.95\linewidth]{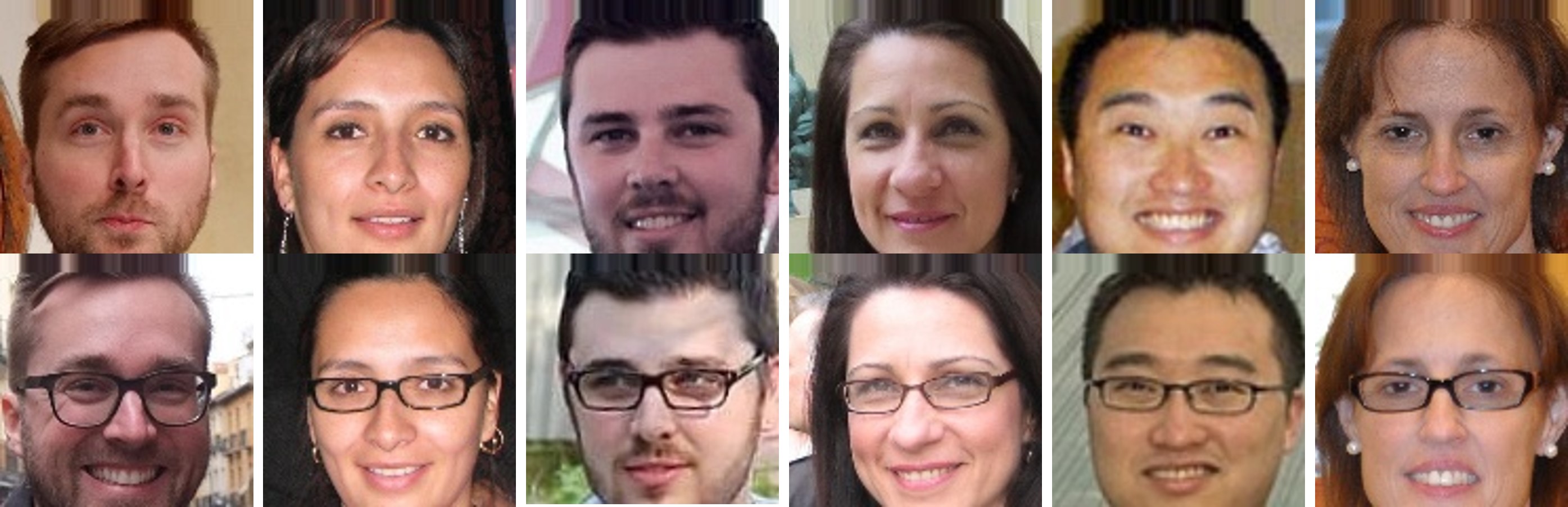}
    \vspace{-0.8em}
    \caption{Sample images of our testing set. For each identity, we show two faces with and without eyeglasses.}
    \label{fig:meglass}
  \end{figure*}

To be consistent with the evaluation protocol (in Section~\ref{sec:evaluation_protocol}), we select two faces with eyeglasses and two faces without eyeglasses from each identity to build our testing set and the total number of images is $6,880$. Fig.~\ref{fig:meglass} shows some examples of testing set with and without eyeglasses.

\subsection{Training set}
\label{training_data}

Two types of training set are adopted, one is only the MsCeleb and the other is the mixture of MsCeleb with synthetic MsCeleb-Eyeglass.
Our MsCeleb clean list has $78,765$ identities and $5,001,877$ images, which is slightly modified from~\cite{wulight}. For each image, we synthesize a eyeglass face image using the method proposed in Section~\ref{sec:eyeglass_image_synthesis}. Therefore, there are totally $10,003,754$ images from $78,765$ subjects in the mixture training set. 

\subsection{Evaluation protocol}
\label{sec:evaluation_protocol}
In order to examine the effect of eyeglass on face recognition thoroughly, we propose four testing protocols to evaluate different methods. 
\begin{enumerate}[I)]
\item All gallery and probe images are without eyeglasses. There are two non-eyeglass face images per person in gallery and probe sets, respectively. 
\item All gallery and probe images are with eyeglasses. There are two eyeglasses face images per person in gallery and probe sets, respectively.  
\item All gallery images are without eyeglasses, and all probe images are with eyeglasses. There are two non-eyeglass face images per person in gallery set and there are two eyeglass face images per person in probe set. 
\item Gallery images contain both eyeglass images and non-eyeglass images, so as probe images. There are four face images (including two non-eyeglass and two eyeglass face images) per person for gallery and probe sets. 
\end{enumerate}


\section{Experiments}
\label{sec:experiments}
Firstly, we evaluate the impact of eyeglasses on face recognition. Second, several experiments are conducted to study the effect of synthetic training data and proposed loss. In experiments, two losses including the classification loss A-Softmax and the metric learning based contrastive loss are investigated. Totally there are four deep learning models are trained based on different losses and training sets. Table~\ref{tab:resnet22-exps} lists the four deep face models. For ResNet-22-A, we apply A-Softmax loss to learn the model from original MsCeleb dataset. We then finetune the model on MsCeleb dataset using contrastive loss to obtain ResNet-22-B. We also finetune the ResNet-22-A model on MsCeleb and its synthetic eyeglasses database to obtain ResNet-22-C. The ResNet-22-D is finetuned from base model ResNet-22-A using gradual sampling strategy with the slope coefficient $\lambda$ of 0.00001 and $p_0$ of 0.

\begin{table*}[t]
	\caption{The configuration settings of different models. ResNet-22-B, ResNet-22-C and ResNet-22-D are all finetuned from ResNet-22-A. GS indicates gradual sampling.}
    \centering
    \begin{tabular}{lcccc}
        \noalign{\smallskip}\hline\noalign{\smallskip}
            \textbf{Model} & \textbf{Training Data} & \textbf{Loss} & \textbf{Strategy} \\
            \noalign{\smallskip}\hline\noalign{\smallskip}
            ResNet-22-A & MsCeleb & A-Softmax~\cite{liu2017sphereface} & - \\
            ResNet-22-B & MsCeleb & Mining-Contrasive & Finetune \\
            ResNet-22-C & Mixture & Mining-Contrasive & Finetune \\
            ResNet-22-D & Mixture & Mining-Contrasive & Finetune+GS \\
        \noalign{\smallskip}\hline
    \end{tabular}
    \label{tab:resnet22-exps}
\end{table*}

\subsection{Experiments settings}
Our experiments are based Caffe~\cite{jiay} framework and Tesla M40 GPU.
All face images are resized to size 120 $\times$ 120, then being normalized by subtracting 127.5 and being divided by 128.
We use SGD with a mini-batch size of 128 to optimize the network, with the weight decay of 0.0005 and momentum of 0.9.
Based on these configurations, the training speed can reach about 260 images per second on single GPU and the inference speed is about 1.5ms per face image.

\subsection{Effect of eyeglass on face recognition}
\begin{figure*}[t]
    \centering
    \includegraphics[width=0.99\linewidth]{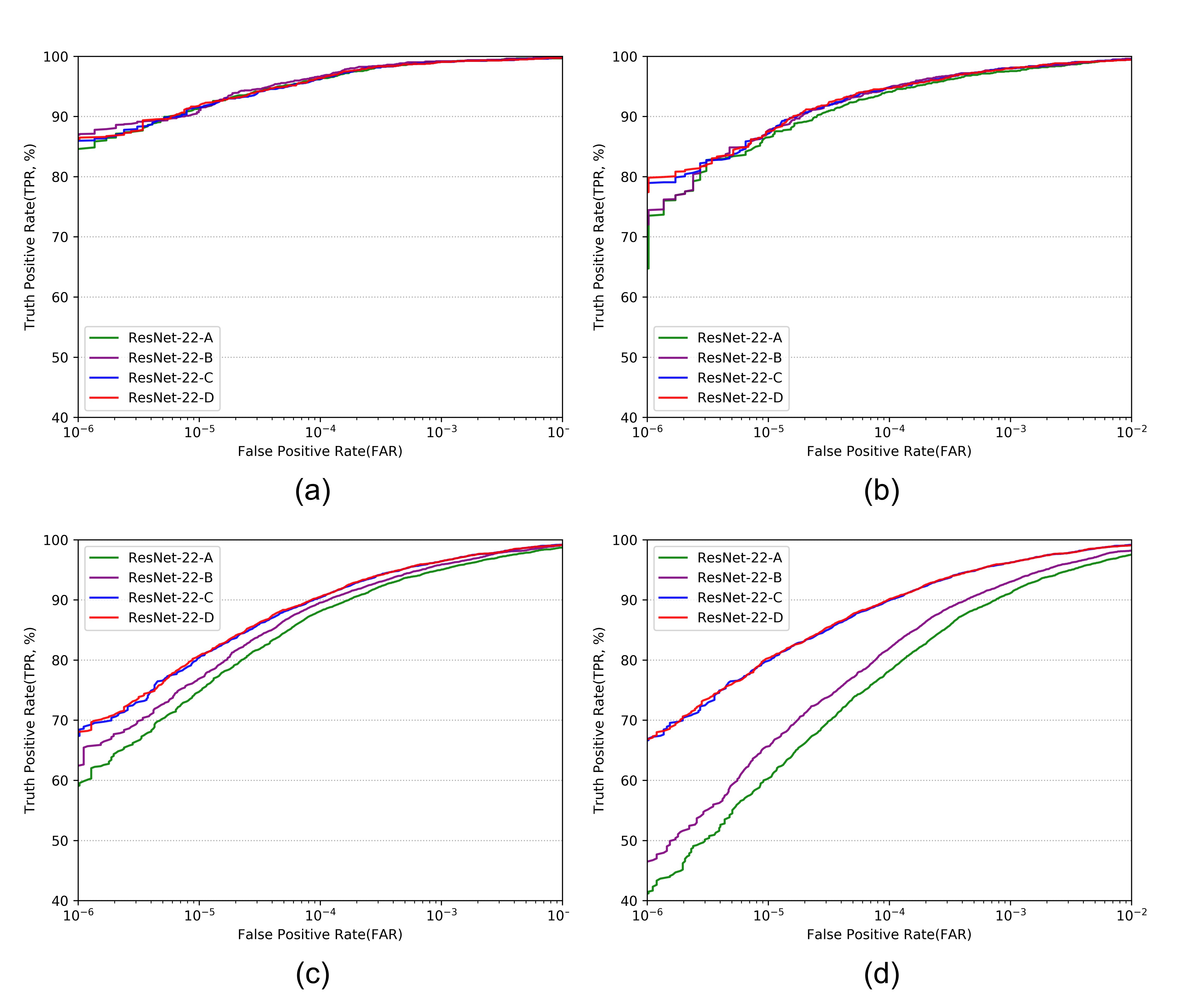}
    \vspace{-0.8em}
    \caption{From (a) to (d): ROC curves of protocols I-IV. For protocols I-II, the curves are almost the same. While for protocols III-IV, ResNet-22-C and ResNet-D models outperform the other two. Especially in protocol IV, they outperform by a large margin (better view on electronic version).}
    \label{fig:roc_curve}
\end{figure*}

In this experiment, we use the original MsCeleb database only as the training set to examine the robustness of traditional deep learning model to eyeglasses. 

\begin{table*}[t]
    \caption{Recognition performance (\%) of ResNet-22-A following protocols I-IV.} 
    \centering
    \begin{tabular}{lcccc}
        \noalign{\smallskip}\hline\noalign{\smallskip}
            Protocol & TPR@FAR=$10^{-4}$ & TPR@FAR=$10^{-5}$ & TPR@FAR=$10^{-6}$ & Rank1 \\
            \noalign{\smallskip}\hline\noalign{\smallskip}
            \rom{1} & 96.14 & 91.49 & 84.68 & 98.48 \\
			\rom{2} & 94.09 & 86.55 & 69.21 & 96.90 \\
			\rom{3} & 88.13 & 74.72 & 59.34 & 95.61 \\
		    \rom{4} & 78.17 & 60.25 & 41.36 & 92.31 \\
        \noalign{\smallskip}\hline
    \end{tabular}
    \label{tab:resnet22_A}
    
\end{table*}

Table~\ref{tab:resnet22_A} shows the results of ResNet-A model tested on four protocols. From the results, one can see that the ResNet-A model achieves high recognition accuracy on protocol I, which is without eyeglass occlusion. However, its performance degrades significantly on protocols II-IV, where eyeglasses occlusion occurs in gallery or probe set, especially for the TPR at low FAR. It indicates that the performance of deep learning model is sensitive to eyeglasses occlusion. 

\subsection{Effectiveness of synthetic data and proposed loss}

\begin{table*}[thp]
    \caption{Recognition performance (\%) of ResNet-22-B following protocols I-IV.}
    \centering
    \begin{tabular}{lcccc}
        \noalign{\smallskip}\hline\noalign{\smallskip}
            Protocol & TPR@FAR=$10^{-4}$ & TPR@FAR=$10^{-5}$ & TPR@FAR=$10^{-6}$ & Rank1 \\
            \noalign{\smallskip}\hline\noalign{\smallskip}
            \rom{1} & 96.61 & 91.26 & 87.02 & 98.60 \\
            \rom{2} & 94.91 & 87.87 & 73.27 & 97.08 \\
            \rom{3} & 89.55 & 76.86 & 62.56 & 96.02 \\
            \rom{4} & 81.96 & 65.68 & 46.71 & 94.18 \\
        \noalign{\smallskip}\hline
    \end{tabular}
    \label{tab:resnet22_B}
\end{table*}

\begin{table*}[thp]
    \caption{Recognition performance (\%) of ResNet-22-C following protocols I-IV.}
    \centering
    \begin{tabular}{lcccc}
        \noalign{\smallskip}\hline\noalign{\smallskip}
            Protocol & TPR@FAR=$10^{-4}$ & TPR@FAR=$10^{-5}$ & TPR@FAR=$10^{-6}$ & Rank1 \\
            \noalign{\smallskip}\hline\noalign{\smallskip}
            \rom{1} & 96.20 & 91.58 & 85.94 & 98.19 \\
            \rom{2} & 94.80 & 87.31 & 78.89 & 96.73 \\
            \rom{3} & 90.35 & 80.40 & 67.93 & 96.67 \\
            \rom{4} & 89.94 & 79.88 & 66.82 & 96.67 \\
        \noalign{\smallskip}\hline
    \end{tabular}
    \label{tab:resnet22_C}
\end{table*}


For comparison, we further train deep face model, ResNet-C, using the mixture of the original MsCeleb and its synthesized eyeglasses version MsCeleb-Eyeglass. Table~\ref{tab:resnet22_B} and Table~\ref{tab:resnet22_C} show the comparison results of ResNet-B and ResNet-C following four protocols. It can be seen that using our synthesized eyeglass face images, it significantly improves the face recognition performance following protocol III-IV, especially at low FAR. It enhances about 20 percent when FAR=$10^{-6}$ on protocol IV, which is the hardest case in four configurations, indicating the effectiveness of virtual face synthesis data for the robustness improvement of face deep model. Moreover, with the face synthesis data , the proposed loss function with gradual sampling, model ResNet-22-D achieves the best results on four protocols.



\begin{table*}[thp]
    \caption{Recognition performance (\%) of ResNet-22-D following protocols I-IV.}
    \centering
    \begin{tabular}{lcccc}
        \noalign{\smallskip}\hline\noalign{\smallskip}
            Protocol & TPR@FAR=$10^{-4}$ & TPR@FAR=$10^{-5}$ & TPR@FAR=$10^{-6}$ & Rank1 \\
            \noalign{\smallskip}\hline\noalign{\smallskip}
            \rom{1} & 96.37 & 91.99 & 86.37 & 98.30 \\
            \rom{2} & 94.68 & 87.54 & 78.68 & 96.78 \\
            \rom{3} & 90.54 & 80.71 & 68.10 & 96.75 \\
            \rom{4} & 90.14 & 80.32 & 66.92 & 96.73 \\
        \noalign{\smallskip}\hline
    \end{tabular}
    \label{tab:resnet22_D}
\end{table*}

Finally, we also plot the ROC curves for four protocols in Fig.~\ref{fig:roc_curve} to further validate the effectiveness of our synthetic training dataset and proposed loss function.

\section{Conclusion}
\label{sec:conclusion}
In this paper, we propose a novel framework to improve the robustness of face recognition with eyeglasses.
We synthesize face images with eyeglasses as training data based on 3D face reconstruction and propose a novel loss function to address this eyeglass robustness problem.
Experiment results demonstrate that our proposed framework is rather effective. In future works, the virtual-synthesis method may be extended to alleviate the impact of other factors on the robustness of face recognition encountered in real life application.
\section{Acknowledgments}
\label{sec:acknowledgments}
This work was supported by the Chinese National Natural Science Foundation Projects \#61473291, \#61572536, \#61572501, \#61573356, the National Key Research and Development Plan (Grant No.2016YFC0801002), and AuthenMetric R\&D Funds.


\end{document}